\documentclass[journal]{IEEEtran}
\usepackage{amssymb}
\usepackage[linesnumbered,ruled,vlined]{algorithm2e}
\SetKwInput{KwInput}{Input}                % Set the Input
\SetKwInput{KwOutput}{Output}              % set the Output
\usepackage{mathrsfs}
\usepackage{multirow}
\usepackage{float}
\usepackage{amssymb}
\usepackage{subfigure} 
\usepackage[utf8]{inputenc}
\usepackage[small]{caption}
\usepackage{graphicx}
\usepackage{amsmath}
\usepackage{amsthm}
\usepackage{booktabs}

\usepackage{pifont}
\usepackage{soul}
\usepackage{url}
\usepackage{color}
\usepackage[table]{xcolor}

\usepackage{hyperref}
\hypersetup{
    colorlinks=true,
    linkcolor=red,
    filecolor=magenta,      
    urlcolor=cyan,
}

\usepackage{cite}
\ifCLASSINFOpdf
\else
\fi
\begin{document}
\title{CIM: Class-Irrelevant Mapping for Few-Shot Classification}

\author{Shuai~Shao$^\dagger$,
        Lei~Xing$^\dagger$,
        Yixin~Chen,
        Yan-Jiang~Wang,
        Bao-Di~Liu$^*$,~\IEEEmembership{Member,~IEEE}
        and~Yicong~Zhou,~\IEEEmembership{Senior Member,~IEEE}
        % and~Jane~Doe,~\IEEEmembership{Life~Fellow,~IEEE}% <-this % stops a space
\thanks{Shao~Shuai, Yan-Jiang~Wang, and Bao-Di~Liu are with the College of Control Science and Engineering, China University of Petroleum (East China), China. 
Lei~Xing is with the College of Oceanography and Space Informatics, China University of Petroleum (East China), China.
Yixin~Chen is with the BIOMIND, China.
Yicong~Zhou is with University of Macau.
}
\thanks{$^\dagger$Shao~Shuai and Lei~Xing are co-first authors. }
\thanks{$^*$Bao-Di~Liu (Email: thu.liubaodi@gmail.com) is corresponding author.}}% <-this % stops a space
% \thanks{Manuscript received April 19, 2005; revised August 26, 2015.}}


\markboth{Journal of \LaTeX\ Class Files,~Vol.~14, No.~8, August~2015}%
{Shell \MakeLowercase{\textit{et al.}}: Bare Demo of IEEEtran.cls for IEEE Journals}



\maketitle

\begin{abstract}
Few-shot classification (FSC) is one of the most concerned hot issues in recent years.
The general setting consists of two phases: (1) Pre-train a feature extraction model (FEM) with base data (has large amounts of labeled samples). (2) Use the FEM to extract the features of novel data (with few labeled samples and totally different categories from base data), then classify them with the to-be-designed classifier. 
The adaptability of pre-trained FEM to novel data determines the accuracy of novel features, thereby affecting the final classification performances.
To this end, how to appraise the pre-trained FEM is the most crucial focus in the FSC community.
It sounds like traditional Class Activate Mapping (CAM) based methods can achieve this by overlaying weighted feature maps.
However, due to the particularity of FSC (e.g., there is no backpropagation when using the pre-trained FEM to extract novel features), we cannot activate the feature map with the novel classes.
To address this challenge, we propose a simple, flexible method, dubbed as \textbf{Class-Irrelevant Mapping (CIM)}. 
Specifically, first, we introduce dictionary learning theory and view the channels of the feature map as the bases in a dictionary.
Then we utilize the feature map to fit the feature vector of an image to achieve the corresponding channel weights. 
Finally, we overlap the weighted feature map for visualization to appraise the ability of pre-trained FEM on novel data.
For fair use of CIM in evaluating different models, we propose a new measurement index, called \textbf{Feature Localization Accuracy (FLA)}.
In experiments, we first compare our CIM with CAM in regular tasks and achieve outstanding performances. Next, we use our CIM to appraise several classical FSC frameworks without considering the classification results and discuss them.

\end{abstract}

% \begin{IEEEkeywords}
% Few-Shot Classification (FSC), Sample-Feature-Mismatch problem, Multi-View Discriminative Semantic Feature (MVDSF), Original-Shift-Feature (OSF), Discriminative-Semantic-Feature (DSF)
% \end{IEEEkeywords}

% \fi
\IEEEpeerreviewmaketitle

\begin{figure}[t]
	\begin{center}
		\includegraphics[width=1.0\linewidth]{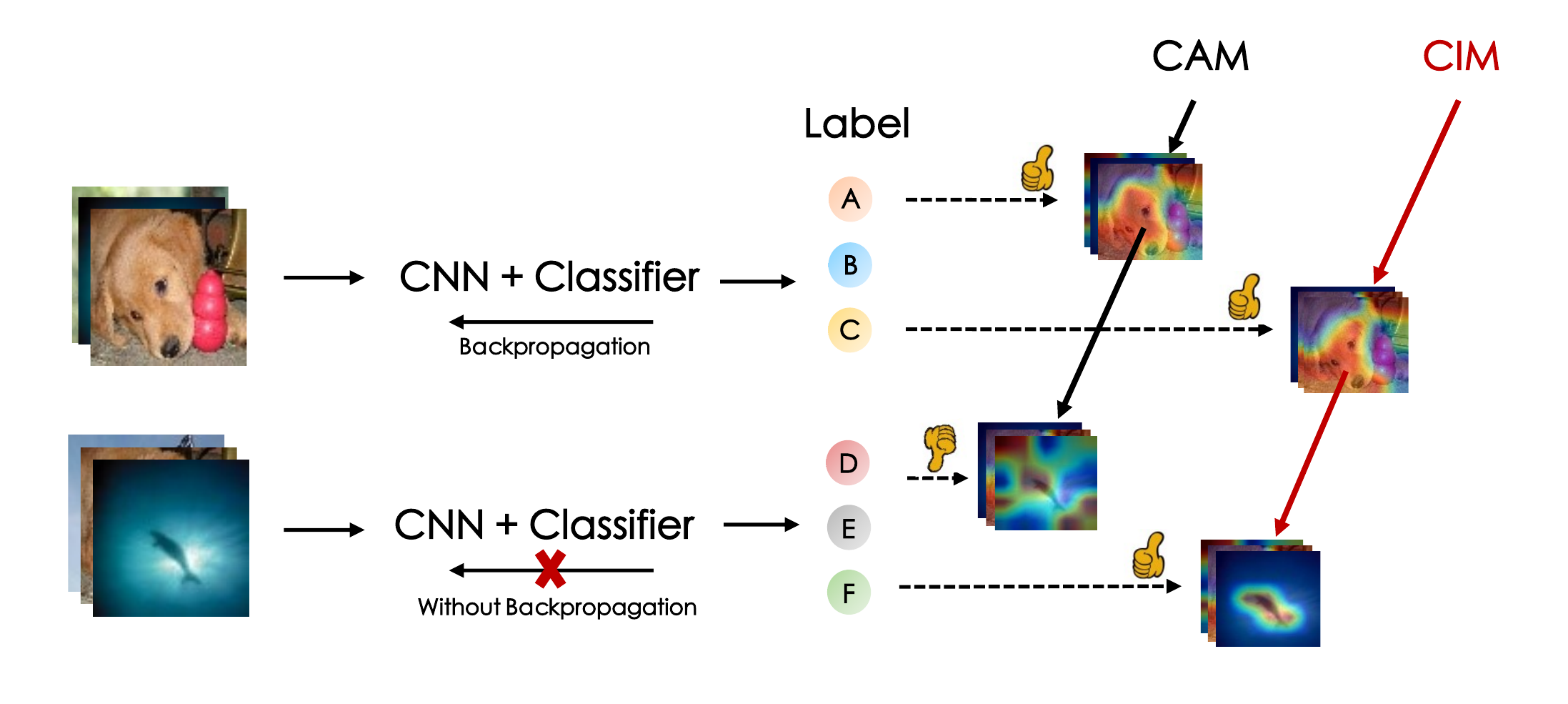}
	\end{center}
    	\caption{
    	Comparison of Class Activate Mapping (CAM) and Class-Irrelevant Mapping (CIM).
    	If the classification model is end-to-end (top), both CAM and CIM achieve satisfactory visualization results.
    	Otherwise (bottom, e.g., employ the pre-trained CNN to extract the features of novel classes, then use traditional classifier, such as SVM, to recognize samples), the novel classes can not activate feature maps for visualization due to the lack of backpropagation. Therefore, CAM is unsuitable for the bottom, but our CIM is unaffected and still obtains beneficial visualization results.
    	}
	\label{fig: CIM_VS_CAM}
\end{figure}

\begin{figure*}[t]
	\begin{center}
		\includegraphics[width=1.0\linewidth]{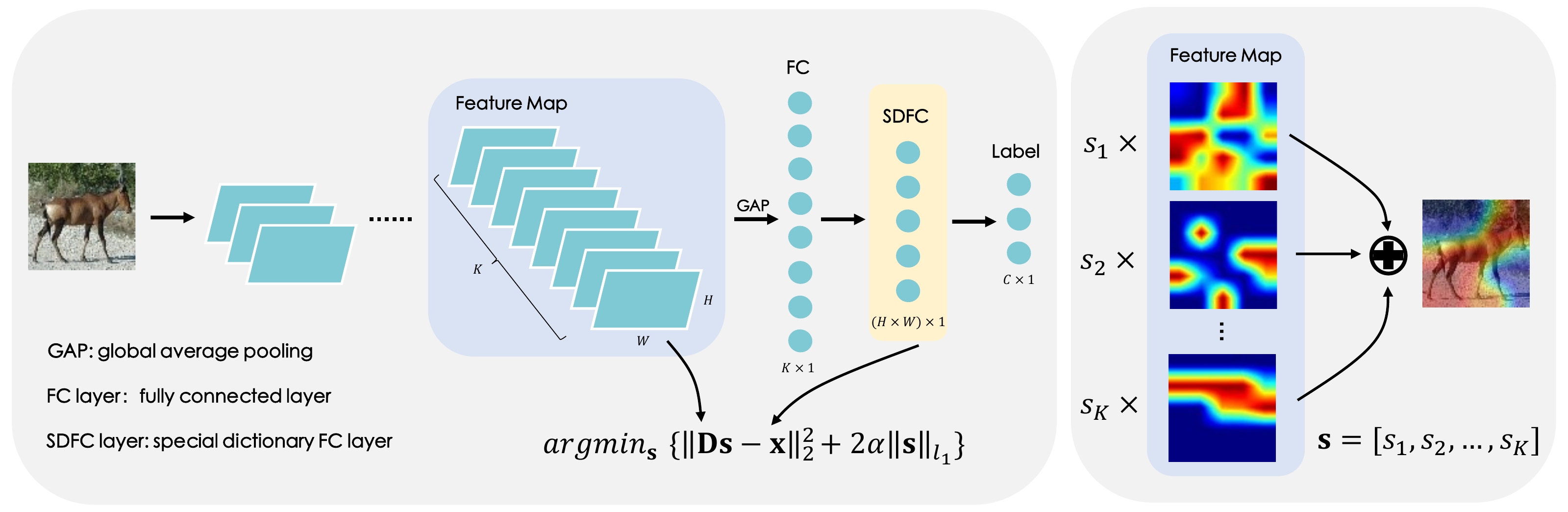}
	\end{center}
    	\caption{
    	The framework to generate Class-Irrelevant Mapping (CIM). 
    	$K$, $W$, and $H$ denote the channel number, the width, and the height of the feature map, respectively. 
    	Our approach designs a specific dictionary FC (SDFC) layer, whose dimensions equal to $H \times W$.
    	After that, the feature map is used as a dictionary $\mathbf{D}$ to fit this particular vector $\mathbf{x}$ (based on SDFC), and obtain the coefficient $\mathbf{s}$.
    	Unlike CAM-based methods, our CIM does not rely on class activation. 
    	So when using this kind of network (has the SDFC layer) to extract novel class features, we can directly obtain the visualization results. 
    % 	Using this kind of network (including the SDFC layer) to extract novel class features still works effectively as a visualization tool.
    % 	Different from traditional CAM based methods, our CIM does not rely on class activation, so when using this kind of network (include SDFC layer) to extract novel class features, it can still be used as a visualization tool
    	}
	\label{fig: Flowcharts}
\end{figure*}

\section{Introduction}
% 关于可解释的机器学习，深度模型的可视化，一直都是研究的热点，早期基于salience map，后来zhou提出基于feature map的可视化模型，用来衡量模型的好坏，从此掀开了新的篇章。谁谁谁都分别做出了杰出的贡献。
% 这些方法的本质，都是为相应层的feature map赋予权重信息，而这些权重信息，都是基于类别激活的，需要依靠网络的反馈来更新。
% 而这种硬性需求，对于某些特殊的任务是不适用的，比如说，小样本学习。

% Shao et.al. MHFC指出，小样本任务里有个基础的问题阻碍它的发展，无论我们pre-train的model在base上表现有多好，都很难适用于新类，（由于跨域的影响），因此得到的新类的特征往往不能精确的描述样本从而给分类造成困扰。
% 因此，评价一个预训练的模型是否在新类上适用，是小样本社区关注的重要问题。
The great strides in deep learning have enabled unprecedented breakthroughs in image classification tasks. The success is partially attributed to a large number of labeled data. However, annotating data is expensive or infeasible, leading to a novel hot issue -- Few-Shot Classification (FSC).
FSC targets to help machines achieve or even surpass human beings' level with scarce labeled samples. 
A general FSC setting includes two components: 
(1) Pre-train. Employ the base data (with large amounts of labeled samples) to train a convolutional neural network (CNN) based Feature Extraction Model (FEM). 
(2) Meta-test. Extract the features of novel data (with only a few labeled samples and totally different categories from base data) by using the trained FEM, then design a classifier (such as support vector machine, logistic regression) to predict their labels.

\cite{dvornik2020selecting} pointed out that a fundamental problem prevents the development of FSC, e.g., regardless of how well the pre-trained FEM performs on base data, it will be enormously challenging to fit into the novel data due to the limitation of \textit{Cross-Domain}. 
The extracted novel features can not describe the novel samples accurately, which causes trouble for the classification.
Therefore, how to evaluate whether a pre-trained FEM is applicable in novel classes is an essential issue of concern to the FSC community.

%The great strides on deep learning has enabled unprecedented breakthroughs in image classification tasks. The success partially attributed to a large number of labeled data. However, annotating data is expensive or even infeasible, which leads to a novel hot issue -- Few-Shot Classification (FSC).
%FSC targets to help machines achieve or even surpass human beings' level with scarce labeled samples. 
%A general FSC setting includes two components: 
%(1) Pre-train. Employ the base data (with large amounts of labeled samples) to train a convolutional neural network (CNN) based Feature Extraction Model (FEM). 
%(2) Meta-test. Extract the features of novel data (has only few labeled samples, and totally different categories from base data) by using the trained FEM, then design classifier (such as support vector machine, logistic regression) to predict their labels.

%\cite{dvornik2020selecting} pointed out that a fundamental problem prevents the development of FSC, e.g., 
%regardless of how well the pre-trained FEM performs on base data, it will be enormously difficult to fit into the novel data due to the limitation of \textit{Cross-Domain}. 
%The extracted novel features can not describe the novel samples accurately, which causes trouble to the classification.
%Therefore, how to evaluate whether a pre-trained FEM is applicable in novel class is an important issue of concern to FSC community.

% 小样本分类的最终预测结果，当然是一个很重要的衡量指标，但是在小样本分类任务里，影响最终分类的因素有很多，最常见的两种关键因素就是模型提取特征的好坏与分类器的设置，也许一个不好的特征配上一个很强的分类器，仍然可以得到不错的结果，因此只用分类精度来评价一个现有模型的好坏，对于评价一个模型来说，是片面的。同时也缺少对模型的可解释性。
Of course, the final classification result is a crucial measurement index. But there exist multiple factors that influence the classification performance, such as the qualities of pre-trained FEM and the to-be-designed classifier. Perhaps a weak FEM with a strong classifier will still yield high classification accuracy. Thus, it is one-sided to evaluate an existing FEM only by the classification accuracy. Besides, this strategy lacks interpretability to the model. 
% The final classification result is of course a crucial measure. But there exist multiple factors influence the result, such as the qualities of pre-trained FEM and the to-be-designed classifier. Perhaps a weak FEM with a strong classifier will still yield good results. Thus, it is one-sided to evaluate an existing FEM only by the classification accuracy. Besides, this strategy lacks interpretability to the model. 

% 因此研究者们尝试通过新类的特征图可视化的方法来对预训练模型进行评价。
% 但在实际应用中却遇到这样一个问题：
% 在通常的小样本设置中，meta-test阶段是没有网络反馈的过程，也就是说，我们无法用新类完成特征图的激活过程从而得到相应的权重，这与CAM based 方法基本原理是相悖的。这听起来似乎只需要我们对网络进行finetune，让网络重新进行反馈，就可以解决这个问题，但是实际上，由于有标签样本数的限制，fine-tune对于小样本分类结果的影响很小，maml已经证明这点，因此绝大部分现有的小样本模型工组，在meta-test阶段，都是without fine-tune的。种种原因导致，现有的CAM方法无法适用于大部分小样本的模型结构。我们的示意图说明了这点。
Inspired by the development of Explainable Deep Learning, researchers attempt to appraise the pre-trained FEM by visualizing the feature map of novel samples by employing Class Activation Map (CAM) based methods. 
CAM was first proposed by \cite{zhou2016learning}. Following, Grad-CAM \cite{selvaraju2017grad}, Grad-CAM++ \cite{chattopadhay2018grad}, Score-CAM \cite{wang2020score} \textit{et.al.} contributed a lot for this community. 
In essence, all these visualization methods attempt to overlay weighted \textit{feature map} to model the salient part of an image. 
They design different structures to calculate channel weights, but all of them are based on the class (to which the sample belongs) activation. That is, they rely on the feedback of the deep neural network to update the weights.
However, in most of existed FSC frameworks, there is no backpropagation process in the meta-test phase. It means that we cannot activate the feature map with the novel class to obtain the corresponding weights, contrary to the basic principle of CAM-based methods. We illustrate an example in \textbf{Figure \ref{fig: CIM_VS_CAM}}.

It sounds like all we need to do is to fine-tune the FEM and drive it to generate class-activated weights for the feature map to tackle this problem. 
However, as the scarce of labeled novel data (as an example, for the typically $1$-shot and $5$-shot case, each category only has $1$ or $5$ labeled samples), fine-tuning only has a limited promotion for FSC (has been proved in \cite{finn2017model}) and is abandoned by the vast majority of FSC approaches.
To this end, the existing CAM-based methods cannot be well applied to the FSC models. 

To address this challenge, we propose \textbf{Class-Irrelevant Mapping (CIM)} to explain and appraise the ability of pre-trained FEM on a novel set.
The essence of a feature map is the description of an image on the same level (e.g., CNN layer) but different focus, which correlates with dictionary learning theory \cite{mallat1993matching} (e.g., a channel of feature map is similar to a base of dictionary).

Inspired by this find, we first design a flexible specific dictionary fully connected layer (SDFC layer) and make the dimension equal to the product of the width and height of the to-be-visualized feature map. Then, we use the feature map as a reconstructed dictionary to fit the feature vector of an image (e.g., the SDFC layer) and introduce a sparse constraint for calculating channel weights. 
Finally, we overlap the weighted feature map to visualize the pre-trained model on novel data and appraise it. We illustrate the flowchart in \textbf{Figure \ref{fig: Flowcharts}}. 
In addition, to quantify metrics for visualization, we propose new evaluation criteria as \textbf{Feature Localization Accuracy (FLA)}, please refer to \textbf{Figure \ref{fig: example_FLA}}.
Notably, our CIM not only performs well on the FSC task, but also is suitable for regular tasks.

% 很灵活，传统的CAM的方法，往往都需要设计专门的网络结构，我们的只需要加入额外的层然后finetune
% FLA

% 我们提出了评价方法并对其评价，可以归纳为两点
% 仅从模型提取新类特征的角度来讲，我们认为ICI>META
% 我们从另一个视角，验证了这个文章中的观点，不是网络越深越好 

In summary, the main contributions focus on:
\begin{itemize}
\item 
Based on the characteristics of the few-shot classification (FSC) architecture, we propose a \textbf{Class-Irrelevant Mapping (CIM)} method, which can be used to measure the quality of the pre-trained feature extraction model (FEM) on novel data. 
\item 
In order to better apply our CIM in the FSC community, we propose a visual quantitative metric, called \textbf{Feature Localization Accuracy (FLA)}.
\item 
Our CIM refers to the dictionary learning theory to calculate channel weights by fitting the feature vector of an image. 
It is a simple, flexible method, which is adequate for most of the existed classification tasks.
\item
In experiments, we first compare our CIM with CAM to evaluate the efficiency in regular classification tasks, then use CIM to appraise several current classical FSC frameworks.
Note that, we merely evaluate the model from the perspective of whether it fits into a new class or not. 
Our purpose in designing CIM is to help researchers in the FSC community to construct more robust and effective networks.

\end{itemize}

\section{Related Work}

\subsection{Few-Shot Classification}
In recent years, Few-shot classification (FSC) has attracted a lot of attention. 
As described above, a high-quality pre-trained feature extraction model (FEM) is crucial to promote the final performances. Therefore, researchers focus on developing novel technologies for designing FEMs. 
ICI \cite{wang2020instance}, CovaMNet \cite{li2019distribution} \textit{et.al.} referred to the standard classification model to design the FEM.
MetaOpt \cite{lee2019meta}, Feat \cite{ye2020few} \textit{et.al.} introduced meta-learning strategy to split the base data to different tasks for pre-training.
GFL \cite{yao2020graph}, KGTN \cite{chen2020knowledge} \textit{et.al.} considered graph structure when constructing the FEM. 
S2M2 \cite{mangla2020charting}, MHFC \cite{shao2021mhfc} \textit{et.al.} fused self-supervised methods to strengthen the FEM.
RFS \cite{tian2020rethinking}, IER \cite{rizve2021exploring} integrated knowledge distillation technology to optimize the FEMs.

\subsection{Class Activation Mapping}
Recently, why deep learning achieves outstanding performance has attracted a lot of attention, which leads to a novel hot issue -- Explainable Deep Learning. 
There exist two perspectives to explain the network. The first one is dubbed as Salience Map, which focused on the pixel of original images, and many related works were proposed, such as \cite{zeiler2014visualizing,simonyan2014deep,smilkov2017smoothgrad,kapishnikov2021guided}.
In the year $2016$, zhou \emph{et al.} proposed the explainable deep learning from another perspective, i.e., Class Activation Map (CAM) \cite{zhou2016learning}.
Following, some improved methods were proposed, including \cite{selvaraju2017grad,chattopadhay2018grad,wang2020score,ramaswamy2020ablation}.
All these methods have various structures, but one indispensable point is the \textit{class activation}. However, this hard need leads to a limitation for some special tasks, such as Few-Shot Classification (FSC).
%Recent years, why the deep learning achieves outstanding performance has attracted a lot of attentions, that leads to a novel hot issue -- Explainable Deep Learning. 
%There exist two perspectives to explain the network. The first one is dubbed as Salience Map, which focused on the pixel of original images, and many related works were proposed, such as \cite{zeiler2014visualizing,simonyan2014deep,smilkov2017smoothgrad,kapishnikov2021guided}.
%In 2016, the first work from another perspective was proposed, e.g., Class Activation Map (CAM) \cite{zhou2016learning}.
%Following, some improved methods were proposed, including \cite{selvaraju2017grad,chattopadhay2018grad,wang2020score,ramaswamy2020ablation}.
%All these methods have different structures, but one indispensable point is the \textit{class activation}. this hard need leads to a limitation for some special tasks, such as Few-Shot Classification (FSC).

\subsection{Dictionary Learning}
Dictionary learning is one of the most popular machine learning theories, which was first proposed by \cite{mallat1993matching}. It is capable of mapping samples' original feature embeddings to the dictionary space. Using the reconstructive dictionary bases to represent samples is helpful to reduce the redundant information of samples.
Dictionary learning was widely applied in various fields, such as image classification \cite{shao2020label,wang2020class}, person re-identification \cite{li2018discriminative,li2019top}, image denoising \cite{zhu2020structured,gong2020low}, etc.
In this paper, we apply it to visualize and explain the deep models.

\begin{figure*}[t]
	\begin{center}
		\includegraphics[width=0.95\linewidth]{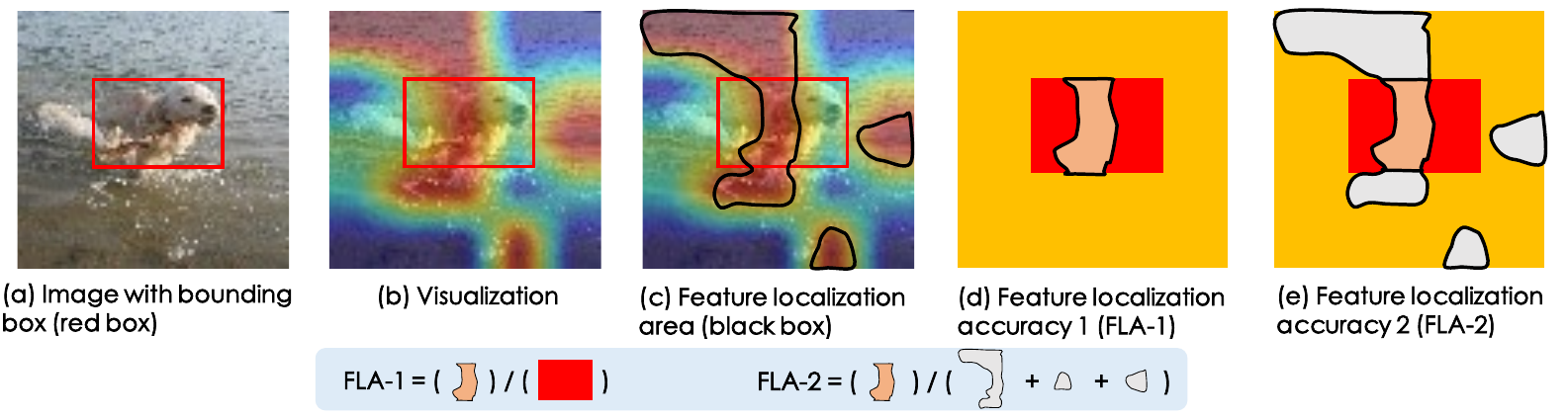}
	\end{center}
    	\caption{
    	An example to introduce the evaluation criterion, e.g., Feature Localization Accuracy (FLA). }
	\label{fig: example_FLA}
\end{figure*}

\section{Methodology}
In this section, we first describe our Class-Irrelevant Mapping (CIM) in detail; then introduce the optimization method for the mapping weight; finally, show the details of how to employ CIM to evaluate FSC models.

\subsection{Class-Irrelevant Mapping}
\label{sec: CIM}

This section shows the details of our Class Irrelevant Mapping (CIM). We view the feature maps as the reconstructive dictionary to model the raw sample.
%Assume that we have an image, input the image into a convolutional neural network $\mathcal{M}$, and obtain the feature map on the $l_{th}$ layer, which is denoted as $f_k(u,v)$, where $k=1,2,\cdots,K$ represents the $k_{th}$ channel, $(u,v)$ indicates the spatial location. 
Take an image as an example, we input the image into a convolutional neural network $\mathcal{M}$, and obtain the feature map on the $l_{th}$ layer, which is denoted as $f_k(u,v)$, where $k=1,2,\cdots,K$ represents the $k_{th}$ channel, $(u,v)$ indicates the spatial location. 
Then pull each channel of feature map to a vector, and construct a matrix as $\mathbf{D} = [\mathbf{d}_1, \mathbf{d}_2, \cdots, \mathbf{d}_K] \in \mathbb{R}^{(H \times W) \times K}$, where $H$, $W$, $K$ denote the height, width, and channel of the feature map, 
an example is shown in \textbf{Figure \ref{fig: Flowcharts}}; $\mathbf{d}_k \in \mathbb{R}^{(H \times W) \times 1}$ indicates the $k_{th}$ 
vector in $\mathbf{D}$. 

The feature map can be seen as the subspace representation of an image, which is similar to the principle of dictionary learning. Each channel corresponds to a base vector of the dictionary.
To obtain reasonable and robust weights independent of the category of the sample, we use the feature map as a reconstructed dictionary to fit the feature vector of the sample, which is defined as $\mathbf{x} \in \mathbb{R}^{dim \times 1}$.
In our task, if we use the feature map on the $l_{th}$ layer to complete the visualization operation, we need to guarantee the length of this vector equal to the product of the height and width of the feature map on the $l_{th}$ layer (e.g., $dim = H \times W$).
Therefore, we need to adjust the network structure to add an additional fully connected (FC) layer with $H \times W$ neurons, which is dubbed as specific dictionary FC (SDFC) layer. The position is shown in \textbf{Figure \ref{fig: Flowcharts}}. 
%Actually, we do not expect this layer to have a large effect on the outcome, as it would disfavor our objective judgment of an already existing FSC network.
Actually, we do not expect this layer to significantly affect the outcome, as it would disfavor our objective judgment of an already existing FSC network.
Fortunately, our ablation study has demonstrated that adding this layer will generate little difference in network performance.

Following, we formulate our objective function to calculate the weights as:

\begin{equation}
\begin{split}
        \mathop {\arg \min}\limits_{\mathbf{s}} 
        f_1(\mathbf{s})
        = \left\| \mathbf{x} - \mathbf{D} \mathbf{s} \right\|_2^2
        + 2\alpha \left \| \mathbf{s} \right\|_{\ell_1}\\
        {\kern 15pt} {\rm{s}}.t. {\kern 3pt} s_k  \ge 0 {\kern 5pt} (k=1,2,\cdots,K)\\
\end{split}
\label{eqa: weight_obj}
\end{equation}
where 
$\left\| \cdot \right\|_2$, $\left\| \cdot \right\|_{\ell_1}$ represent $\left( \cdot \right)$'s $\ell_2$-norm, $\ell_1$-norm. 
$\alpha$ is used to adjust sparsity of $s$.
$\mathbf{x} \in \mathbb{R}^{dim \times 1}$, denotes the feature vector of the sample.
$\mathbf{D} \in \mathbb{R}^{dim \times K}$, indicates the one layer's feature map of the sample.
$dim = H \times W$.
$\mathbf{s}=[s_1,s_2,\cdots,s_K] \in \mathbb{R}^{K \times1}$ is the to-be-learned weight vector. $s_k \,\, (k=1,2,\cdots,K)$ indicates the element in $\mathbf{s}$, corresponds to the weight of $k_{th}$ channel of feature map (the optimization process, please refer to Section \textbf{ Optimization}). 
Thus, we define the objective function of our class-irrelevant mapping as:

\begin{equation}
\begin{split}
    \mathcal{M}(u,v) = \sum_{k=1}^K s_k f_k(u,v) 
\end{split}
\label{eqa: map_obj}
\end{equation}
where $\mathcal{M}(u,v)$ reflects the key spatial grid.
% where $\mathcal{M}(u,v)$ is activated by a special feature vector

% indicates the crucial spatial grid $(u,v)$ activated by dictionary learning.

% is capable of indicating the key point of spatial grid $(u,v)$.

\subsection{Optimization for Mapping Weight}
\label{sec: optimization}
Observe \textbf{Equation \ref{eqa: weight_obj}}, $\mathbf{D}$ is a constant matrix. We introduce the Alternating Direction Method of Multipliers (ADMM) \cite{boyd2011distributed} to solve the optimization problem.

\textbf{(1)} Introduce an auxiliary variable ${\bf{q}}$ and reformulate \textbf{Equation \ref{eqa: weight_obj}} as:

\begin{equation}
\begin{split}
        \mathop {\arg \min}\limits_{\mathbf{s},\mathbf{q}} f_2(\mathbf{s},\mathbf{q}) =
        \left\| \mathbf{x} - \mathbf{D}\mathbf{s} \right\|_2^2 + 2\alpha \left\| \mathbf{q} \right\|_{\ell_1}\\
        {\kern 5pt} s.t. {\kern 2pt} \mathbf{s} = \mathbf{q}, {\kern 3pt} s_k  \ge 0 {\kern 5pt} (k=1,2,\cdots,K)
\end{split}
\label{eqa: ADMM1}
\end{equation}

\textbf{(2)} Introduce the Lagrangian operator and rewrite \textbf{Equation \ref{eqa: ADMM1}} as:

\begin{equation}
\begin{split}
        &\mathop {\arg \min}\limits_{\mathbf{s},\mathbf{q},\mathbf{m}} f_3(\mathbf{s},\mathbf{q},\mathbf{m}) \\
        &= \left\| \mathbf{x} - \mathbf{D}\mathbf{s} \right\|_2^2 
        + 2\alpha \left\| \mathbf{q} \right\|_{\ell_1}
        + \left<\mathbf{m},  \mathbf{s} - \mathbf{q} \right> 
        + \rho \left\| \mathbf{s} - \mathbf{q} \right\|_2^2
\end{split}
\label{eqa: ADMM2}
\end{equation}
where $\mathbf{m}$ is the augmented lagrangian multiplier, and $\rho>0$ denotes the penalty parameter. 

\textbf{(3)} Alternative update $\mathbf{s}$, $\mathbf{q}$, $\mathbf{m}$ until \textbf{Equation \ref{eqa: ADMM2}} convergence, the solutions are formulated as follows:

\begin{equation}
\begin{split}
        \left\{\begin{array}{lll}
            \mathbf{s} 
            = \left( \mathbf{D}^T \mathbf{D} + \rho \mathbf{I} \right)^{-1}
            \left( \mathbf{D}^T \mathbf{x} + \rho {\kern 2pt} \mathbf{q} - \mathbf{m} \right)\\
            \mathbf{q} 
            = max \left\{ \mathbf{s} + \frac{1}{{\rho}} \mathbf{m} - \frac{\alpha}{{\rho}} \mathbf{1}, \mathbf{0} \right\}\\
            \mathbf{m} 
            = \mathbf{m} + \theta \left( \mathbf{s} - \mathbf{q} \right)  
        \end{array}\right.
\end{split}
\label{equation: ADMM_solutions}
\end{equation}
where $\mathbf{I}$ denotes the identity matrix; $\mathbf{1}$ and $\mathbf{0}$ indicate the one vector and zero vector; $\theta$ denotes the gradient degree.

% ############################### compare_CAM_CIM_visualization begin
\begin{figure*}[t]
	\begin{center}
		\includegraphics[width=1\linewidth]{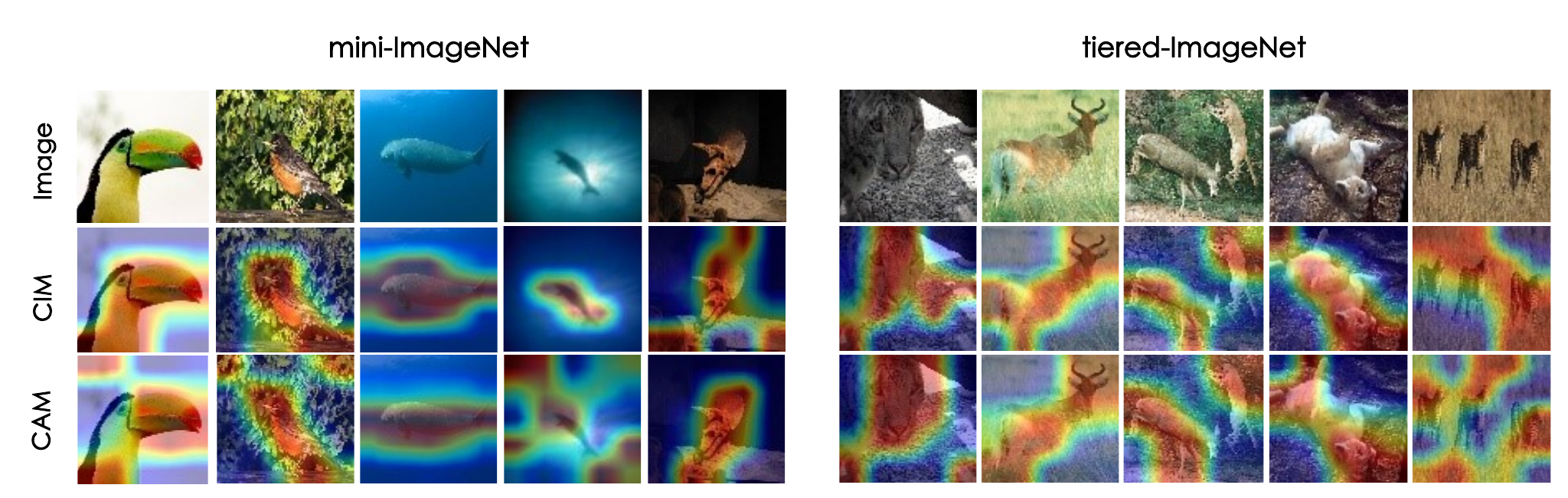}
	\end{center}
    	\caption{
    	Visualization of CIM and CAM. The images come from \textbf{base set} of mini-ImageNet and tiered-ImageNet.}
	\label{fig: compare_CAM_CIM_visualization}
\end{figure*}
% ############################### compare_CAM_CIM_visualization end

% ############################### FLA_camvscim begin
\begin{figure}[h]
    \centering
    % \subfigure[]{
    \begin{minipage}[t]{0.5\linewidth}
    	\begin{center}
    		\includegraphics[width=1\linewidth]{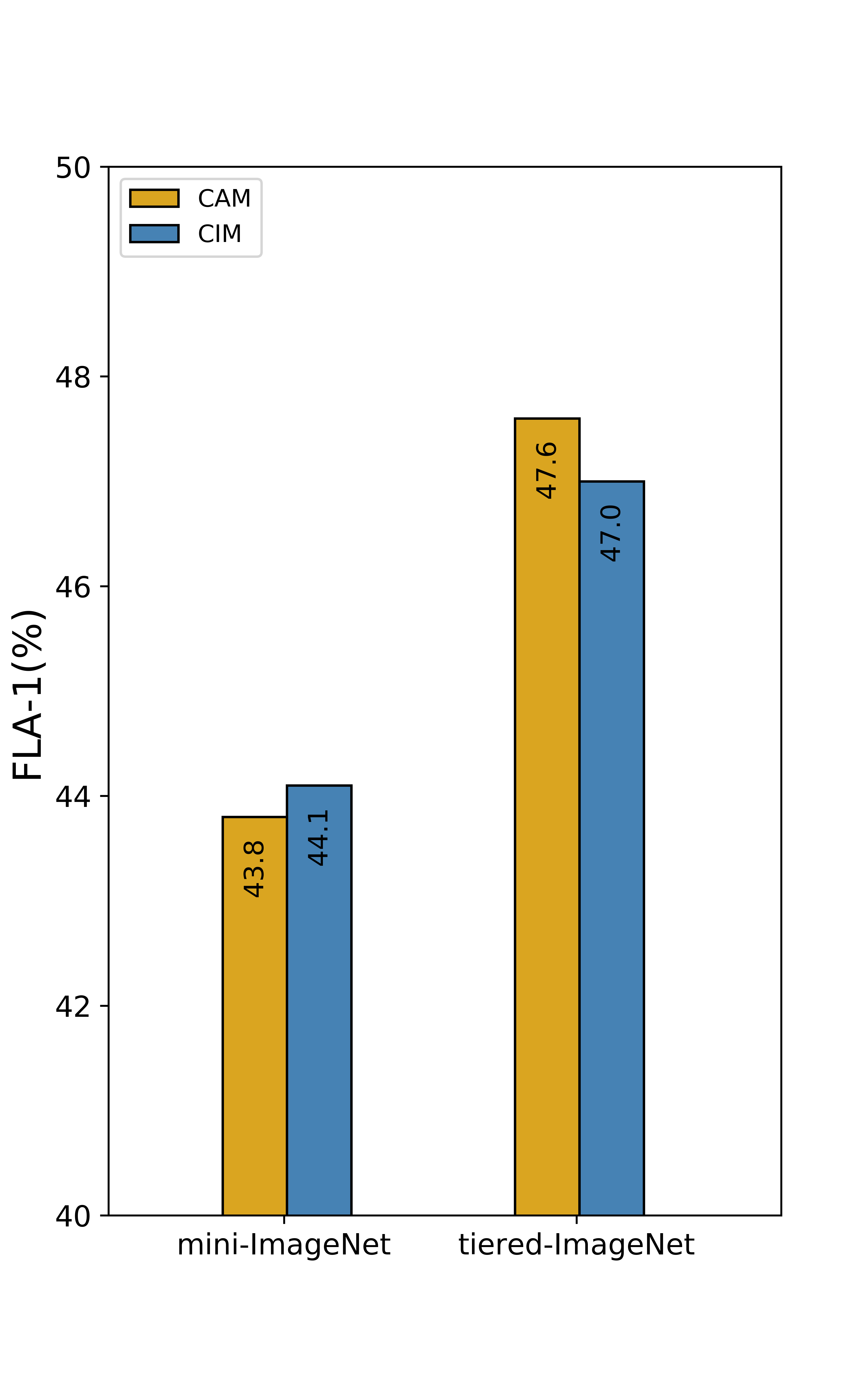}
    	\end{center}
    % 	\caption{}
    	\label{fig: FLA1_camvscim}
    \end{minipage}%
    % }
    % \subfigure[]{
    \begin{minipage}[t]{0.5\linewidth}
    	\begin{center}
    		\includegraphics[width=1\linewidth]{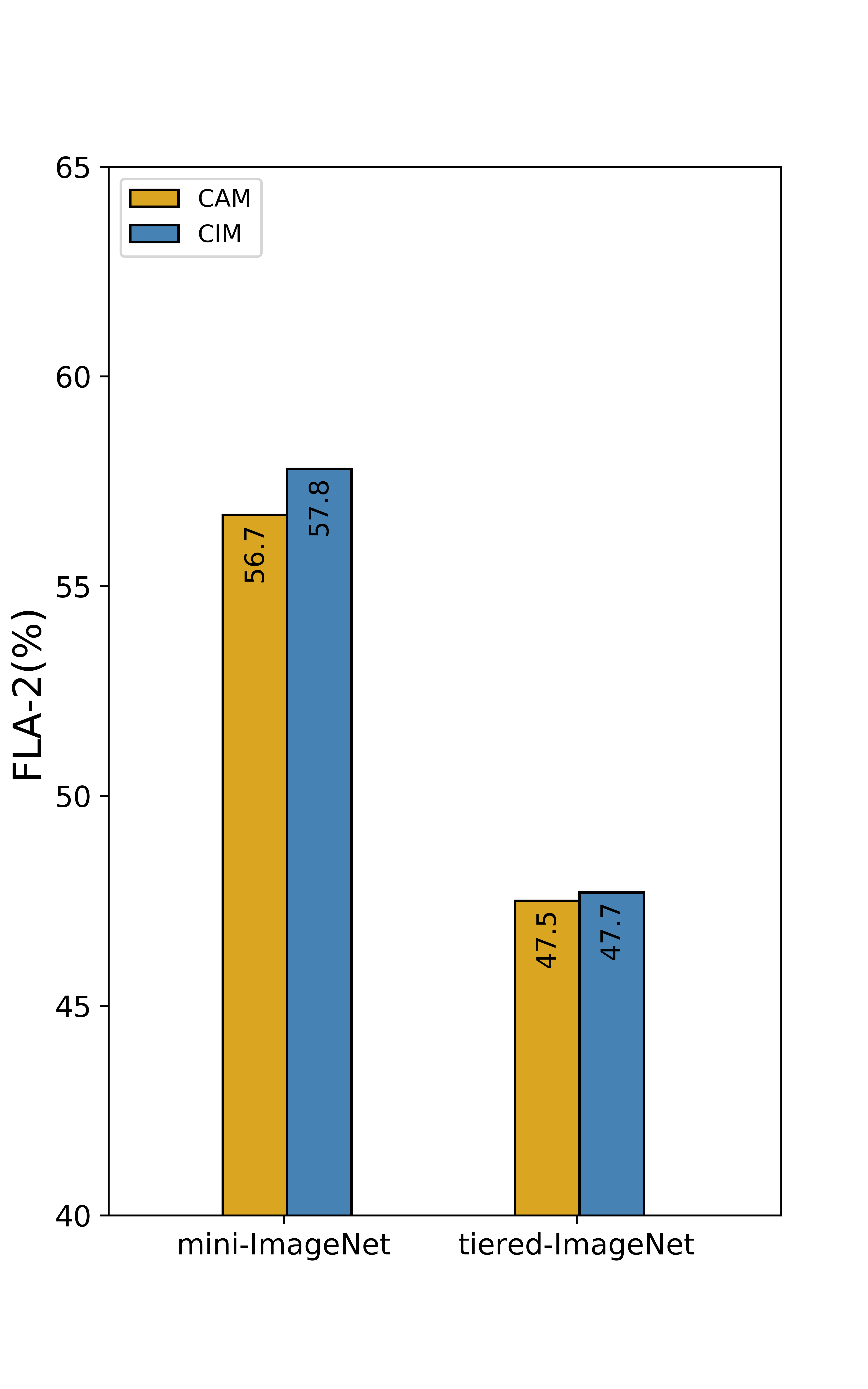}
    	\end{center}
    % 	\caption{}
    	\label{fig: FLA2_camvscim}
    \end{minipage}%
    % }
    \caption{Comparison FLA ($\%$) of our CIM with CAM on mini-ImageNet and tiered-ImageNet. Larger is better.}
    \label{fig: FLA_camvscim}
\end{figure}
% ############################### FLA_camvscim end

% ############################### compare_network_visualization begin
\begin{figure*}[t]
	\begin{center}
		\includegraphics[width=1\linewidth]{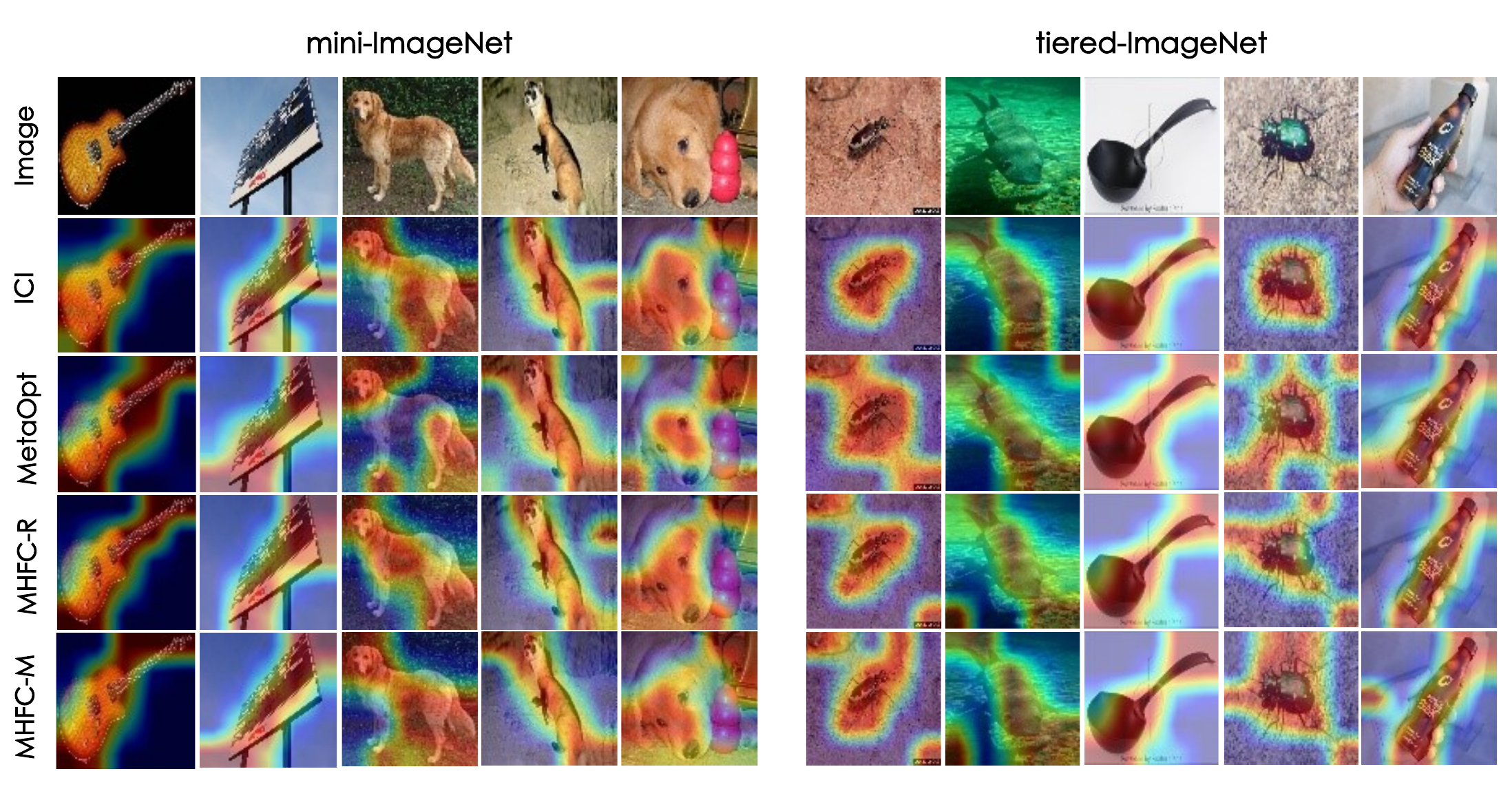}
	\end{center}
    	\caption{
    	Visualization of CIM on different FSC networks. The images come from \textbf{novel set} of mini-ImageNet and tiered-ImageNet.}
	\label{fig: compare_network_visualization}
\end{figure*}
% ############################### compare_network_visualization end

% ############################### alpha begin
\begin{figure}[t]
    \centering
    % \subfigure[]{
    \begin{minipage}[t]{1\linewidth}
    	\begin{center}
    		\includegraphics[width=1.0\linewidth]{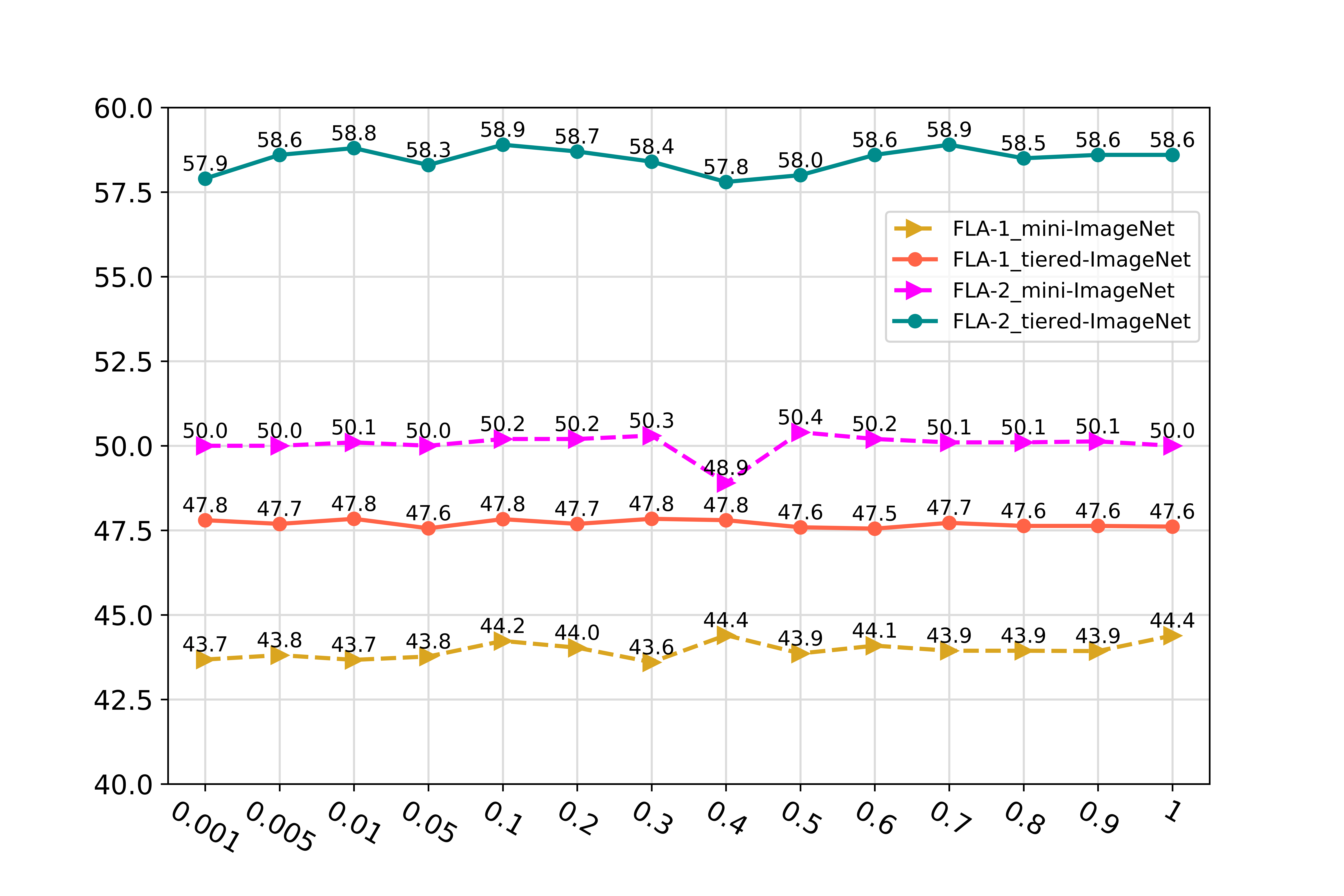}
    	\end{center}
    % 	\caption{}
    	\label{fig: alpha_fla1}
    \end{minipage}%
    % }
    % \subfigure[]{
    % \begin{minipage}[t]{0.5\linewidth}
    % 	\begin{center}
    % 		\includegraphics[width=1\linewidth]{Figures/alpha_FLA2.eps}
    % 	\end{center}
    % % 	\caption{}
    % 	\label{fig: alpha_fla2}
    % \end{minipage}%
    % }
    \caption{Ablation study to evaluate the influence of $\alpha$ (x-axis) on mini-ImageNet and tiered-ImageNet.}
    \label{fig: alpha}
\end{figure}
% ############################### alpha end

\subsection{CIM for Few-Shot Classification}
\subsubsection{Problem Setup of Few-Shot Classification}
The General few-shot classification setting (FSC) includes two stages, e.g., pre-train and meta-test. 
\textbf{(1)} In pre-train stage, we use base set $\mathcal{D}_{base} = \{(x_i,y_i) |{\kern 1pt} y_i \in \mathcal{C}_{base} \}$ to train a standard classification model $\mathcal{A}_{\theta}$, where $x$ and $y$ denote the sample and corresponding label, respectively; 
$\mathcal{C}_{base}$ indicates the base category set;
$\theta$ denotes the parameters of the network. 
\textbf{(2)} In the meta-test stage, we use the trained model $\mathcal{A}_{\theta}$ as the feature extraction model to extract the feature of novel data, then recognize them. The novel data is denoted as $\mathcal{D}_{novel} = \{(x_j,y_j) |{\kern 1pt} y_j \in \mathcal{C}_{novel} \}$, where $\mathcal{C}_{novel}$ denotes the novel category set. Note that, $C_{base} {\kern 2pt} \cap {\kern 2pt} C_{novel} = \emptyset$.

Due to the scarce of labeled novel samples, fine-tuning is not appliable for the few-shot task.
Besides the difference between $\mathcal{C}_{base}$ and $\mathcal{C}_{novel}$, the network parameters $\theta$ cannot be well applied to the novel data, making it impossible to activate feature mapping with novel classes.
Therefore, we use our class-irrelevant mapping (CIM) to model the heat map and reveal if this network is suitable for new classes.

% \subsection{Discussion about Mapping Weight}
% Traditional dictionary learning aims to better represent sample by transforming the raw feature to a dictionary space. 
% Usually, we set different dictionary bases through our experience, and we need to update them in optimization. 
% While in this paper, we employ the feature map to model the dictionary, each channel can be viewed as a dictionary base. 
% Notably, the feature map is fixed so that we don't need to update the

\subsubsection{FSC Network}
% 放到方法里面，写的再详细一点，然后加上一个表格，描述五种模型的不同

In the FSC community, researchers have designed lots of strategies to improve the robustness of the FEM. In this article, four representative model structures are selected to evaluate.
(1) \textbf{ICI} \cite{wang2020instance} designed a standard classification model as the pre-trained model. 
(2) \textbf{Metaopt} \cite{lee2019meta} introduced meta-learning strategy to make the network \textit{learning to learn}. 
% (3) \textbf{Feat} \cite{ye2020few} integrated both traditional classification model and meta-learning based classification model as the pre-trained model.
(3) \textbf{MHFC-R} \cite{shao2021mhfc} introduced rotation based auxiliary loss to a standard classification model, the rotated degrees include $r \in \{0^{\circ}, 90^{\circ}, 180^{\circ}, 270^{\circ} \}$.
(4) \textbf{MHFC-M} \cite{shao2021mhfc} introduced mirror based auxiliary loss to a standard classification model, the mirrored ways consist of $m \in \{vertically, horizontally, diagonally \}$.

% \begin{figure}[t]
% 	\begin{center}
% 		\includegraphics[width=0.9\linewidth]{Figures/example_FLA1.pdf}
% 	\end{center}
%     	\caption{
%     	An example to introduce the evaluation criterion, e.g., Feature Localization Accuracy (FLA).
%     	}
% 	\label{fig: example_FLA}
% \end{figure}

% \begin{figure}[t]
% 	\begin{center}
% 		\includegraphics[width=1.0\linewidth]{Figures/example_FLA2.pdf}
% 	\end{center}
%     	\caption{
%     	An example to introduce the evaluation criterion, e.g., Feature Localization Accuracy (FLA).
%     	}
% 	\label{fig: example_FLA}
% \end{figure}

% \begin{figure*}[t]
% 	\begin{center}
% 		\includegraphics[width=1.0\linewidth]{Figures/example_FLA1.pdf}
% 	\end{center}
%     	\caption{
%     	An example to introduce the evaluation criterion, e.g., Feature Localization Accuracy (FLA).
%     	}
% 	\label{fig: example_FLA}
% \end{figure*}

\subsubsection{Feature Localization Accuracy}

To evaluate the quality of the pre-trained model, we propose a new evaluation criterion as \textbf{Feature Localization Accuracy (FLA)}. It aims to reveal if the pre-trained model pays attention to the crucial and correct visual cues on the novel image. 
Specifically, we first collect bounding boxes for images from \cite{ILSVRC15}, just like \textbf{Figure \ref{fig: example_FLA}(a)}. 
Then we utilise the visualization methods, such as CAM, CIM, to generate hot map through \textbf{Equation \ref{eqa: map_obj}}, see \textbf{Figure \ref{fig: example_FLA}(b)}. The darker the red, the more attention of the network.
Next, we pick the region of the hot map with pixel values larger than $0.6 \times 255$ as the feature localization area, e.g., the area inside the black box in \textbf{Figure \ref{fig: example_FLA}(c)}.
Finally, we define the Feature Localization Accuracy (FLA), which consists of two components, FLA-1 and FLA-2.  
FLA-1 indicates the proportion of information that we focus on within the bounding box, see \textbf{Figure \ref{fig: example_FLA}(d)}.
FLA-2 indicates that the information we focus on in the bounding box accounts for the proportion of all the information we follow with interest, see \textbf{Figure \ref{fig: example_FLA}(e)}. Note that, the larger the two values, the better the model.

\section{Experiments}
In this section, we first introduce the employed datasets, then compare our CIM technology with classical CAM method on the regular task to evaluate the efficiency of CIM, next design ablation studies to analyse the factors that influence our CIM.
Finally, we use our CIM to appraise several existed classical FSC networks and discuss them.
We conduct our experiments on a Tesla-$V100$ GPU with $32G$ memory. 
All the source codes will be made available to the public.  

\subsection{Dataset}
This paper evaluates our CIM on two benchmark datasets of FSC field, e.g., mini-ImageNet \cite{vinyals2016matching}, tiered-ImageNet \cite{ren2018meta}. Both of them are the subsets of ImageNet dataset \cite{russakovsky2015imagenet}. 
\textbf{mini-ImageNet} includes $100$ classes, and we select $64$ classes as the base set, $16$ classes as the validation set, $20$ classes as the novel set. Each class include $600$ images with the size of $84 \times 84$. 
\textbf{tiered-ImageNet} is composed of $608$ classes with $600$ images per class. We select $351$ classes as the base set, $97$ classes as the validation set, $160$ classes as the novel set. All the image size is $84 \times 84$. 
We collect the bounding box from \cite{ILSVRC15}.

% 注意，由于mini和bounding box出处有些不一样，（相同编号不同图片），我们用boundingbox这里的图片替换了原来的mini-imagenet数据集。

\begin{table}[!t]
\caption{Comparison results on mini-ImageNet with or without SDFC layer. Smaller change is better.}
\begin{center}
\begin{tabular}{lcccccc}
\toprule 
% \multicolumn{1}{c}{\multirow{2}{*}{Setting}}
% \multicolumn{1}{l}{\multirow{2}{*}{ \textbf{Setting}}}
\multicolumn{2}{c}{\multirow{2}{*}{\textbf{Method}}}
% &\multicolumn{1}{l}{\multirow{2}{*}{}}
& \multicolumn{2}{c}{\textbf{mini-ImageNet}} 

% & \multicolumn{2}{c}{CUB} 
\\ 
\cmidrule(l){3-4}
% \multicolumn{1}{c}{}
\multicolumn{1}{c}{}
&& \textbf{$5$-way $1$-shot} & \textbf{$5$-way $5$-shot}   \\ 
\midrule
\multirow{2}{*}{\textbf{ICI}}  
&& $56.06 $   & $75.70 $       \\
% & +SDFC  
% & $\rightarrow 56.25$   & $\rightarrow 76.80$       \\
&\cellcolor{gray!30} +SDFC  
&\cellcolor{gray!30} $\rightarrow 56.25$   &\cellcolor{gray!30} $\rightarrow 76.80$       \\
\midrule
\multirow{2}{*}{\textbf{MetaOpt}}  
&& $62.64 $   & $78.63 $       \\
% & +SDFC  
% & $\rightarrow 60.58$   & $\rightarrow 75.06$       \\
&\cellcolor{gray!30} +SDFC  
&\cellcolor{gray!30} $\rightarrow 60.58$   &\cellcolor{gray!30} $\rightarrow 75.06$       \\
\midrule
\multirow{2}{*}{\textbf{MHFC-R}}  
&& $61.49 $   & $80.34 $       \\
% & +SDFC  
% & $\rightarrow 61.50$   & $\rightarrow 80.12$       \\
&\cellcolor{gray!30} +SDFC  
&\cellcolor{gray!30} $\rightarrow 61.50$   &\cellcolor{gray!30} $\rightarrow 80.12$       \\
\midrule
\multirow{2}{*}{\textbf{MHFC-M}}  
&& $61.79 $   & $80.62 $       \\
% & +SDFC  
% & $\rightarrow 60.86$   & $\rightarrow 80.50$       \\
&\cellcolor{gray!30} +SDFC  
&\cellcolor{gray!30} $\rightarrow 60.86$   &\cellcolor{gray!30} $\rightarrow 80.50$       \\
% \midrule
% MetaOpt  
% & $62.64 \rightarrow 60.58$   & $78.63 \rightarrow 75.06$       \\
% %Feat  & $55.50 \rightarrow 55.50$   & $55.50 \rightarrow 55.50$       \\
% MHFC-R  
% & $61.49 \rightarrow 61.50$   & $80.34 \rightarrow 80.12$       \\
% MHFC-M  
% & $61.79 \rightarrow 60.86$   & $80.62 \rightarrow 80.50$       \\
\bottomrule
\end{tabular}
% }
\end{center}
\label{tab: SDFC} 
\end{table}

% \begin{figure}[t]
% 	\begin{center}
% 		\includegraphics[width=1.0\linewidth]{Figures/alpha.eps}
% 	\end{center}
%     	\caption{
%     	Ablation study to evaluate the influence of $\alpha$ (x-axis) on mini-ImageNet and tiered-ImageNet. }
% 	\label{fig: alpha}
% \end{figure}

\begin{table}[t]
\caption{Comparison FLA ($\%$) of different FSC models by using our CIM on \textbf{novel data}. Larger is better.}
\begin{center}
\begin{tabular}{lcccccc}
\toprule 
% \multicolumn{1}{c}{\multirow{2}{*}{Setting}}
% \multicolumn{1}{l}{\multirow{2}{*}{ \textbf{Setting}}}
\multicolumn{1}{l}{\multirow{2}{*}{\textbf{Method}}}

& \multicolumn{2}{c}{\textbf{mini-ImgaeNet}} 
& \multicolumn{2}{c}{\textbf{tiered-ImgaeNet}} 
% & \multicolumn{1}{l}{\multirow{2}{*}{\textbf{Average}}}
% & \multicolumn{2}{c}{CUB} 
\\ 
\cmidrule(l){2-5}
% \multicolumn{1}{c}{}
\multicolumn{1}{c}{}
                        
& \textbf{FLA-1} & \textbf{FLA-2}  & \textbf{FLA-1} & \textbf{FLA-2} \\ 

\midrule
ICI  
& $43.9$          &$58.9$          & $45.8$          & $57.6$       \\
MetaOpt  
& $44.7$          &$57.9$          & $48.0$          & $53.8$       \\
MHFC-R  
& $\bf{45.3}$ &$58.7$          & $\bf{48.2}$ & $\bf{59.6}$     \\
MHFC-M  
& $46.0$          &$\bf{59.1}$ & $47.7$          & $55.7$    \\
\bottomrule
\end{tabular}
% }s
\end{center}
\label{tab: compare_network} 
\end{table}

\subsection{Evaluate Our CIM}

\subsubsection{Compare with CAM}
To evaluate FSC by CIM, it needs to be guaranteed that CIM itself can achieve good results. Therefore, let's start by comparing the performance of CIM with classical CAM \cite{zhou2016learning} in regular tasks. We conduct the experiments on mini-ImageNet and tiered-ImageNet. Specifically, we view the base data of the two datasets as the training data to train a network with a fixed ResNet-12 backbone (e.g., pre-trained ICI network). Then use our proposed Feature Localization Accuracy (FLA) to evaluate the performance. 
The visualization results and FLA based results on \textbf{base data} are separately shown in \textbf{Figure \ref{fig: compare_CAM_CIM_visualization}, \ref{fig: FLA_camvscim}}. 
We can see that, different methods (CIM and CAM) are suitable to different images and they achieve similar performance as a whole. Thus, we can safely use it to evaluate FSC networks.

\subsubsection{Ablation Study}
% Obviously, the additional SDFC has little influence to the original classification performance. 因此我们可以放心使用。
From \textbf{Figure \ref{fig: Flowcharts}}, we know that our CIM is adaptive for all kinds of existed networks if we add an extra specific dictionary FC (SDFC) layer. This step does not require us to retrain a model, just need to fine-tune the model with extremely few resource consumption. 
And next, we have to discuss whether this extra layer will have a large impact on the final FSC results (what we expect is no). 
Therefore, we design ablation studies to look at the influence of SDFC layer. Specifically, we use mini-ImageNet to compare the final classification results with or without it. The results are listed in \textbf{Table \ref{tab: SDFC}}. ICI, MetaOpt, MHFC-R, MHFC-M are the classical few-shot classification networks.
Obviously, we see that the additional SDFC layer has little effect on the network, and further demonstrates the generalizability of our CIM.  

% 这意味着，我们的方法可以做到客观的评价已经存在的模型

% This means that our method can do objective evaluation of models that already exist

% This means that our approach can objectively evaluate existing models

From \textbf{Equation \ref{eqa: weight_obj}}, we know that parameter $\alpha$ maybe affect the performance of our CIM. To evaluate it, we make ablation studies on mini-ImageNet and tiered-ImageNet to see the change of FLA. \textbf{Figure \ref{fig: alpha}} lists the influence. From the results, we conclude that our method is not sensitive to this parameter, and it is helpful to the stability of visualization.

% alpha是控制稀疏度的。评测不同的alpha对FLA的影响，在mini上。

\subsection{New Appraised Criterion for FSC}

\subsubsection{Appraise FSC Network}

The above section has proved that CIM is a good choose to visualize the pre-trained FSC model on novel set. 
In this section, we use our CIM to appraise the existed FSC networks on \textbf{novel data}. All of them are based on the ResNet-12 backbone. Some visualization results are shown in \textbf{Figure \ref{fig: compare_network_visualization}}, and the FLA of all images are listed in \textbf{Table \ref{tab: compare_network}}. We list some observations.
    
(1) We rank the four models from our perspective (rank the sum value of each model), that is, MHFC-R ($211.8$) $\rightarrow$ MHFC-M ($208.5$) $\rightarrow$ ICI ($206.2$) $\rightarrow$ MetaOpt ($204.4$).
(2) ICI based model adopts a regular classification network, while MetaOpt introduces meta-learning strategy.
Compared to the results of ICI and Metaopt, we find that ICI achieves similar performance to Metaopt or even surpasses it. 
It demonstrates that using a meta-learning strategy alone seems to have limited help for FSC in our perspective.
(3) Compared the two self-supervised models (MHFC-R and MHFC-M) with ICI, they both achieve better performances. It demonstrates the efficiency of self-supervised auxiliary loss for the FSC task.
Besides, MHFC-R performs better than MHFC-M, implying that rotation loss may be better suited for FSC than mirror loss.

% ############################### FLA_network begin
\begin{figure}[t]
    \centering
    % \subfigure[]{
    \begin{minipage}[t]{0.5\linewidth}
    	\begin{center}
    		\includegraphics[width=1\linewidth]{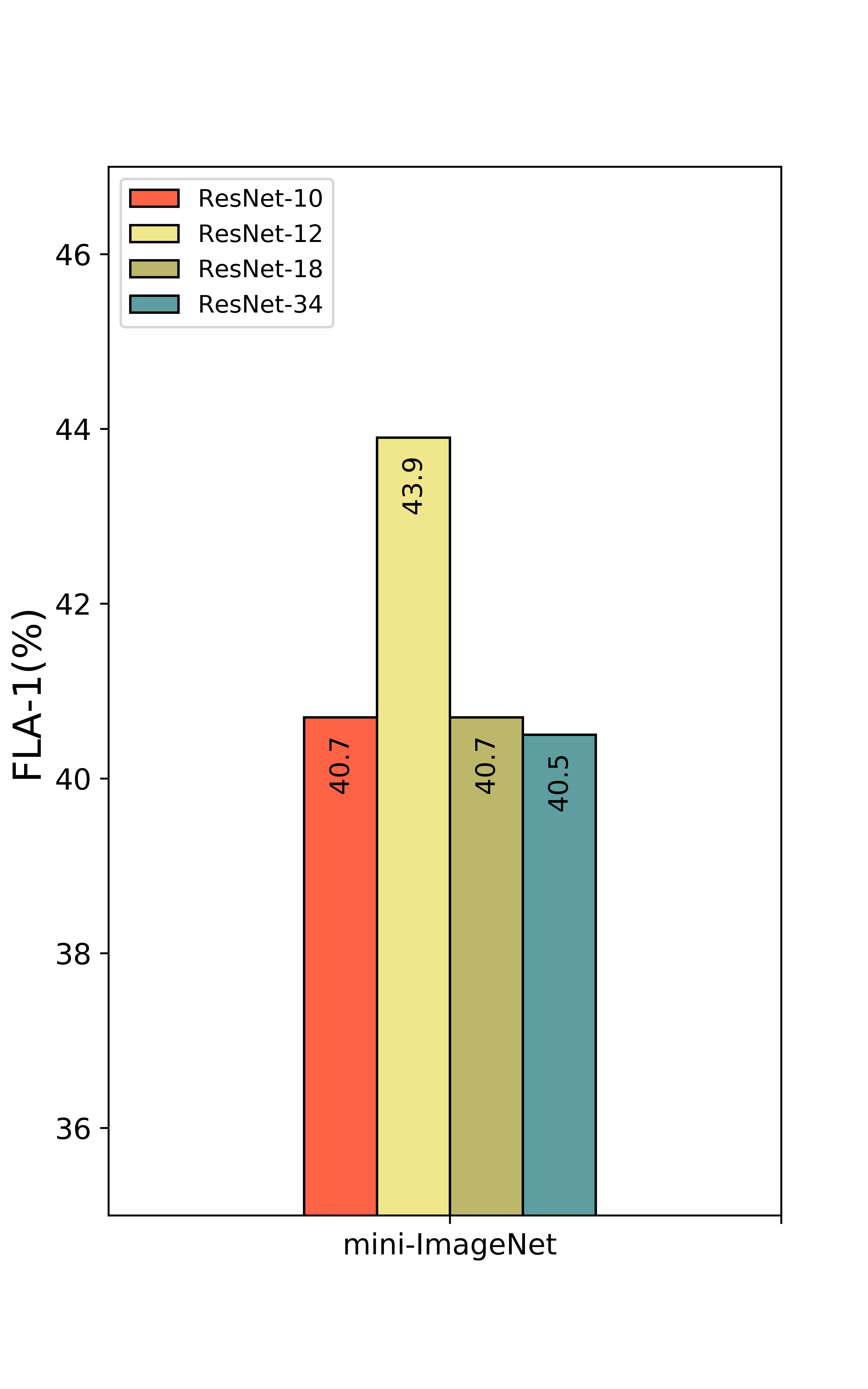}
    	\end{center}
    % 	\caption{}
    	\label{fig: FLA1_network}
    \end{minipage}%
    % }
    % \subfigure[]{
    \begin{minipage}[t]{0.5\linewidth}
    	\begin{center}
    		\includegraphics[width=1\linewidth]{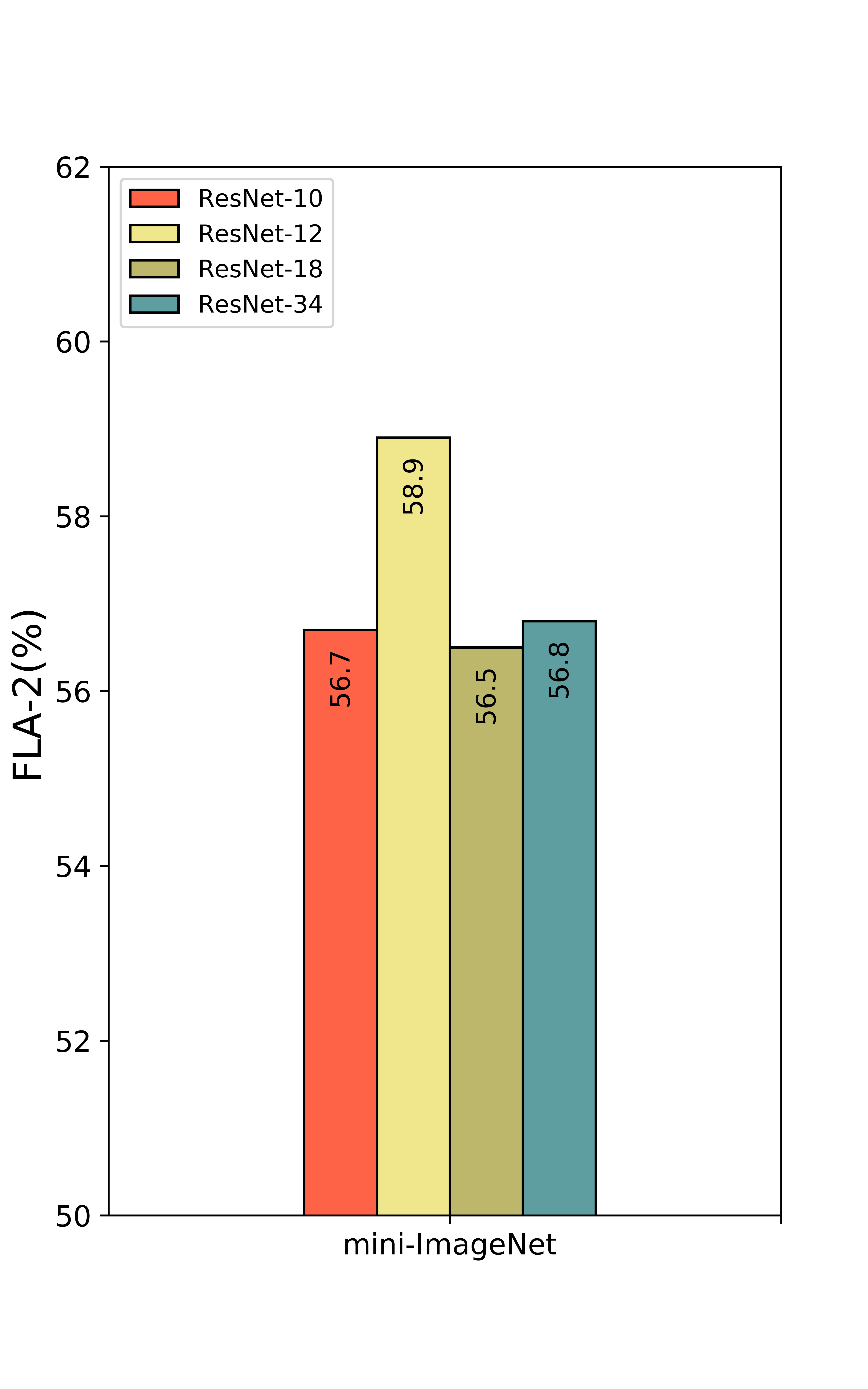}
    	\end{center}
    % 	\caption{}
    	\label{fig: FLA2_network}
    \end{minipage}%
    % }
    \caption{Comparison FLA ($\%$) of our CIM with different backbones on mini-ImageNet. Larger is better.}
    \label{fig: FLA_network}
\end{figure}
% ############################### FLA_network end

% ############################### FLA_baseline_dropout_smooth end
\begin{figure}[t]
    \centering
    % \subfigure[]{
    \begin{minipage}[t]{0.5\linewidth}
    	\begin{center}
    		\includegraphics[width=1\linewidth]{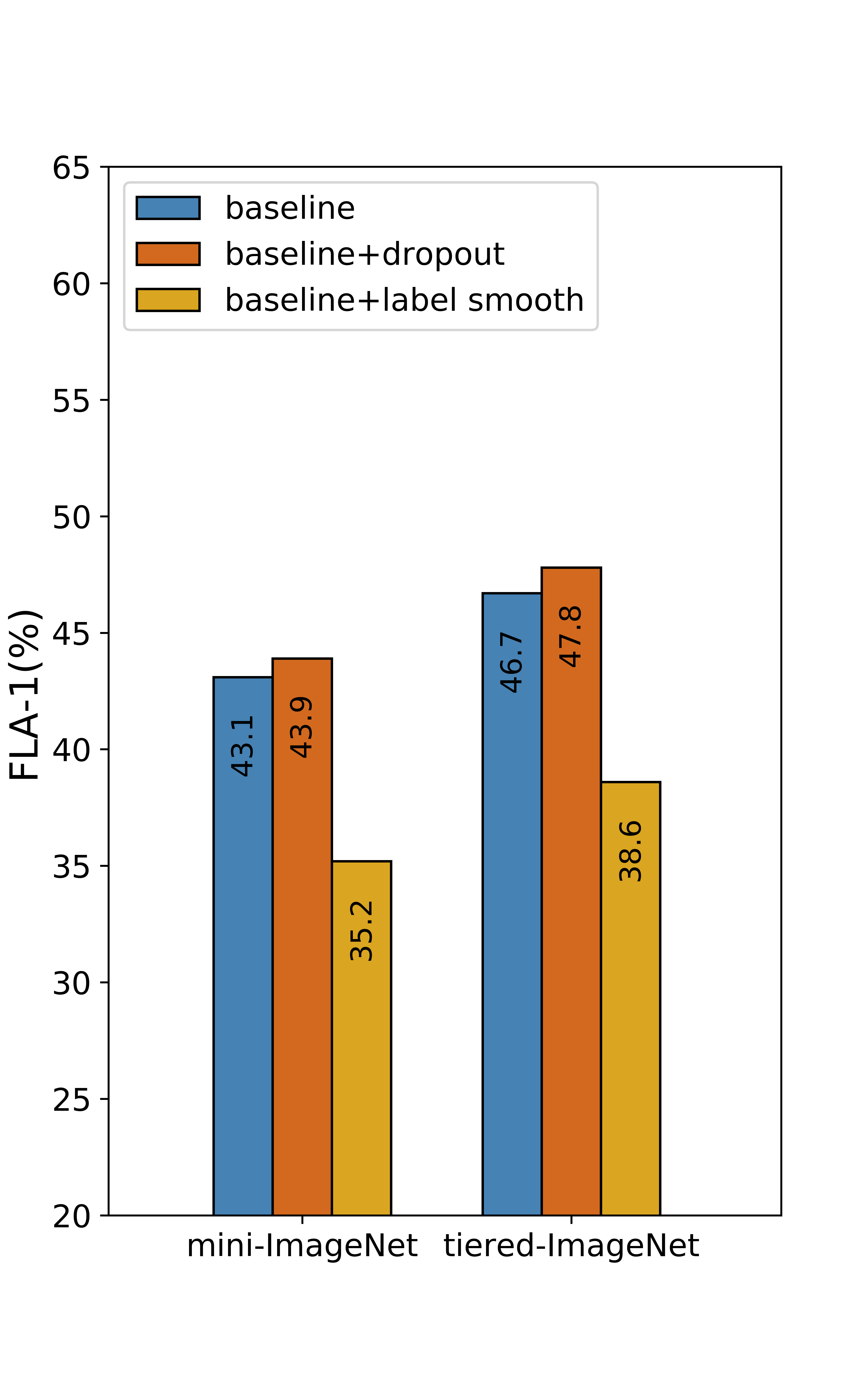}
    	\end{center}
    % 	\caption{}
    	\label{fig: FLA1_baseline_dropout_smooth}
    \end{minipage}%
    % }
    % \subfigure[]{
    \begin{minipage}[t]{0.5\linewidth}
    	\begin{center}
    		\includegraphics[width=1\linewidth]{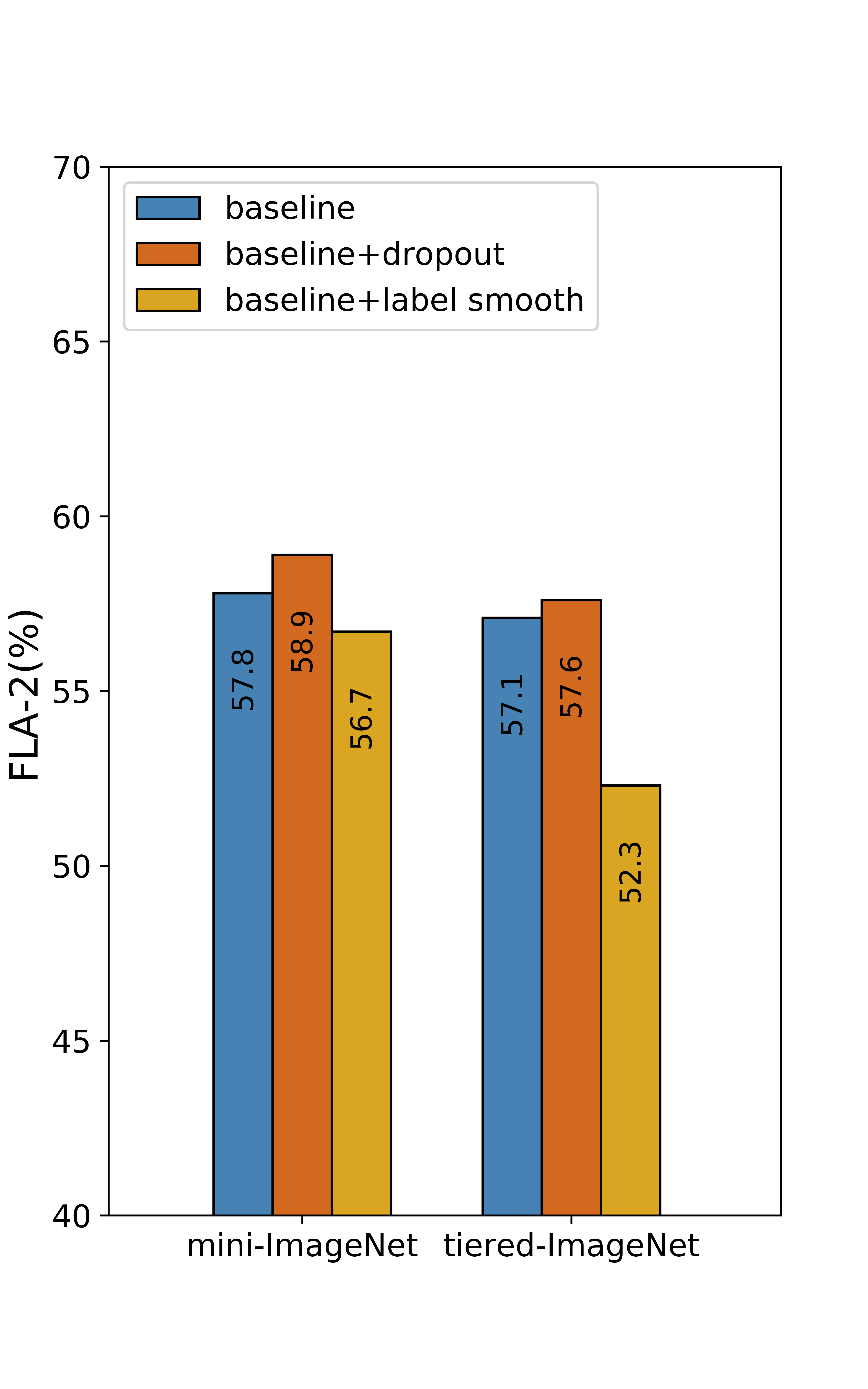}
    	\end{center}
    % 	\caption{}
    	\label{fig: FLA2_baseline_dropout_smooth}
    \end{minipage}%
    % }    
    \caption{Comparison FLA ($\%$) of our CIM with different tricks in the network. Larger is better.}
    \label{fig: FLA_baseline_dropout_smooth}
\end{figure}
% ############################### FLA_baseline_dropout_smooth end

\subsubsection{What Factors Influence the Pre-trained Model?}

In FSC based network, we know that some factors influence the final classification results, such as the backbone, dropout, smoothing loss function.
Thus, it is interesting to see how they influence the pre-trained model on \textbf{novel data} from our CIM-based view.
Specifically, we employ the ICI network as the example to look at the FLA on mini-ImageNet and tiered-ImageNet.

For the backbones, we select four kinds of popular ones (e.g., ResNet-10, ResNet-12, ResNet-18, ResNet-34) to observe the results, which is listed in \textbf{Figure \ref{fig: FLA_network}}. 
We are surprised to find that the structure based on ResNet-12 can achieve much higher performances than other backbones in our evaluation indicators.
Actually, this phenomenon coincides with a view of \cite{chen2019closer}: In this particular area of few-shot classification, the depth of the network can affect the performance of classification, but it is not the case that the deeper the network, the better the outcome.
Or there is another possibility: Our testing network is based on improvements of ICI. It is possible that some parameters (tricks) in the ICI network structure highly match the Resnet-12 architecture.

For the other factors, we list the comparison results in \textbf{Figure \ref{fig: FLA_baseline_dropout_smooth}}. All the results are based on the ResNet-12 backbone. 
As we all know, dropout is not generally used in normal CNN architectures due to the presence of batch normalization. However, in the special field of FSC, adding dropout to the network can help improve performance.
In addition, label smoothing is a common strategy in CNN, which has been used in MetaOpt. But when we introduce it into the generic ResNet architecture, we find that it has a serious negative impact on the results.
Of course, it may just be an accident, just not suitable for the ICI based network, but we hope that researchers are more careful when using this trick when designing FSC model.

\subsection{Discussion}
%In this section, we first verify the validity of CIM, then use CIM to evaluate the FSC structure.
%Some FSC methods may not perform well on our metrics, but the performance of the final classification is good.
%Because we mentioned earlier that there are many factors affect the final classification results, our evaluation criteria are only used to assess whether the pre-trained model can adapt to the novel class, that is, how well it performs on cross-domain with limited labeled samples.
%Our ultimate goal is to help researchers design model structures that are more suitable for FSC tasks.
In this section, we first verify the validity of CIM, then use CIM to evaluate the FSC structure.
Several FSC methods may not perform well on our metrics, but the performance of the final classification is satisfactory.
Because we mentioned earlier that there are many factors affect the final classification results, our evaluation criteria are only used to assess whether the pre-trained model can adapt to the novel class, that is, how well it performs on cross-domain with limited labeled samples.
Our ultimate goal is to help researchers design model structures that are more suitable for FSC tasks.

\section{Conclusion}
%According to the characteristics of FSC task, this paper designs Class Irrelevant Mapping (CIM) as a new measurement method to evaluate FSC framework. 
%It uses the feature maps as dictionary bases to calculate the weights without relying on the class activation. This method is not only suitable for FSC tasks but also suitable for traditional architecture.
%In the experimental section, we evaluate some classical FSC methods through CIM and analyze them.
%In future work, we'd like to introduce graph regularization method to fit dictionary bases and achieve more robust weights for feature map.
According to the characteristics of the FSC task, this paper designs Class Irrelevant Mapping (CIM) as a new measurement method to evaluate the FSC framework. 
It uses the feature maps as dictionary base vectors to calculate the weights without relying on the class activation. This method is not only suitable for FSC tasks but also suitable for traditional architecture.
In the experimental section, we evaluate some classical FSC methods through CIM and analyze them.
In future work, we'd like to introduce the graph regularization method to fit dictionary bases and achieve more robust weights for feature maps.

% or


% \appendices

% % you can choose not to have a title for an appendix
% % if you want by leaving the argument blank
% \section{}

\section*{Acknowledgment}
The paper was supported by 
the National Natural Science Foundation of China (Grant No. 62072468), 
the Natural Science Foundation of Shandong Province, China (Grant No. ZR2019MF073), 
the Fundamental Research Funds for the Central Universities, China University of Petroleum (East China) (Grant No. 20CX05001A),
the Graduate Innovation Project of China University of Petroleum (East China) YCX2021117,
and the Graduate Innovation Project of China University of Petroleum (East China) YCX2021123,
% the Open Research Fund from Shandong Provincial Key Laboratory of Computer Network (No. SDKLCN-2018-01), 
the Qingdao Science and Technology Project (No. 17-1-1-8-jch). 
% the Major Scientific and Technological Projects of CNPC (No. ZD2019-183-008), 
% the Creative Research Team of Young Scholars at Universities in Shandong Province (No.2019KJN019), 

\ifCLASSOPTIONcaptionsoff
  \newpage
\fi



% \bibliographystyle{IEEEtran}

\bibliographystyle{IEEEtran}
\bibliography{IEEEabrv.bib}
\end{document}